\documentclass[runningheads]{llncs}
\usepackage{amsmath}
\usepackage{amssymb}
\usepackage{multirow}
\usepackage{graphicx}
\usepackage[T1]{fontenc}
\usepackage{graphicx}
\usepackage{tabularx}
\newcolumntype{C}{>{\centering\arraybackslash}X}
\newcolumntype{R}{>{\raggedleft\arraybackslash}X}
\newcolumntype{L}{>{\raggedright\arraybackslash}X}

\begin{document}
\title{$F_{\beta}$-plot - a visual tool for evaluating imbalanced data classifiers}
%
%
\author{Szymon Wojciechowski\orcidID{0000-0002-8437-5592} \and \\
Michał Woźniak\orcidID{0000-0003-0146-4205}}
\authorrunning{Szymon Wojciechowski et al.}

%
\institute{Wroclaw University of Science and Technology\\
Wybrzeże Wyspiańskiego 27, 50-370 Wrocław, Poland\\
\email{[michal.wozniak,szymon.wojciechowski]@pwr.edu.pl}}
\maketitle
\begin{abstract}
One of the significant problems associated with imbalanced data classification is the lack of reliable metrics. This runs primarily from the fact that for most real-life (as well as commonly used benchmark) problems, we do not have information from the user on the actual form of the loss function that should be minimized. Although it is pretty common to have metrics indicating the classification quality within each class, for the end user, the analysis of several such metrics is then required, which in practice causes difficulty in interpreting the usefulness of a given classifier. Hence, many aggregate metrics have been proposed or adopted for the imbalanced data classification problem, but there is still no consensus on which should be used. An additional disadvantage is their ambiguity and systematic bias toward one class. Moreover, their use in analyzing experimental results in recognition of those classification models that perform well for the chosen aggregated metrics is burdened with the drawbacks mentioned above. Hence, the paper proposes a simple approach to analyzing the popular parametric metric $F_{\beta}$. We point out that it is possible to indicate for a given pool of analyzed classifiers when a given model should be preferred depending on user requirements.

\end{abstract}
\section{Introduction}
The problem of classifier evaluation has been known and addressed in the literature for years \cite{Japkowicz:2011}. One of the critical issues is the selection of appropriate quality metrics and the proposed experimental protocol. This work will address the first problem for the imbalanced data classification. In this task (we will consider the most popular binary problem), we are dealing with the difference in abundance between classes, specifically referring to the dominant class as the majority class and the less abundant class as the minority class. Furthermore, we typically presume that an error made in the minority class has a greater cost than an error made in the majority class. The difficulty in classifying imbalanced data arises primarily as we lack data on the costs of these errors, known as the loss function.

Usually, two simple metrics that indicate errors made on a given class are analyzed, i.e., $specificity$ or $sensitivity$ (also called $recall$), or $recall$ and $precision$, which indicates how many objects are classified in a minority class. Of course, to say anything about the quality of a given classifier, we need to analyze at least two of these metrics - $recall$ and $precision$, which are usually used nowadays. Unfortunately, we can't determine the best classifier without details about the loss function or the significance of each metric in the user's context.
Additionally, following the common human inclination to communicate information with a single numerical indicator, many aggregated metrics have been suggested, merging information about $precision$ and $recall$, or $specificity$ and $sensitivity$ into an aggregated metric. Unfortunately, most people do not realize that such an approach has serious drawbacks, i.e., such metrics are ambiguous (they do not indicate which values of the simple metrics characterize the evaluated model), and also choosing them as a criterion often leads to the selection of classifiers biased towards the majority class.

Most researchers employ computer experiments to evaluate the predictive performance of proposed models. Usually, the results are promising, and the authors conclude that their classifiers can outperform the state-of-the-art algorithms. However, it should be noted that Wolpert's \emph{no free lunch} theorem concludes that the best learning algorithm does not exist \cite{Wolpert:2001}, so any conclusions drawn from experiments are conditioned on given datasets and a chosen experimental protocol, which specifies, among other things, which metrics are taken into account during the comparison.

Moreover, imbalanced data classification has become an arena for fighting for the best values of selected metrics, which, as we pointed out, can be biased. According to Goodhart's law \cite{Strathern:1997}, "When a measure becomes a target, it ceases to be a good measure".

Unfortunately, finding the right metrics for classifying imbalanced data is very difficult because we generally do not know how costly the errors made on the different classes are. Often, the assumptions made in the experiments about these costs (expressed in the assumed set of metrics) are far from reality, and the authors of the papers do not bother to show comparisons for other cases, i.e., for different costs of the mentioned errors.

Such an approach could be detrimental to the development of classification methods, as only solutions for a specific relatively standard set of metrics, which express particular (not always true) expectations about the quality of classifiers, are preferred and published. This results in solutions that do not perform well for a given set of metrics not being published and thus not being promoted even if they could be helpful in other metrics settings and thus find application in specific practical problems.
Of course, one should not be surprised by this approach, as the primary motivation of most researchers is to publish their work. However, most journals are not interested in negative results because they are not as attractive to readers and do not seem interesting \footnote{Fortunately, more and more people see the value in publishing negative results as well, and an increasing number of journals are choosing to do so (see, e.g., https://doi.org/10.15252/embr.201949775)}. 
This means that even if we design a classifier evaluation experiment according to the guidelines for classifier testing such as \cite{Demsar2006,Garcia:2009,Garcia:2010}, there is still room for metric manipulation \cite{Stapor:2021}.

This work focuses on a proposal to evaluate the classifier quality using the parametric measure $F_{\beta}$, which commonly is chosen for the value $\beta=1$ and does not follow the end-user expectations. 
The criticism of $F1$ or $F-score$ can be addressed since the value of the parameter $\beta$ \emph{de facto} indicates how much more critical $recall$ is compared to $precision$ \cite{Hand:2018}. 

This work contributes to imbalanced data classifier evaluation, especially it points out the problem of inadequate comparison of classification methods, which, through an improper choice of metrics, promotes models with characteristics suitable for only particular user preferences, i.e., a specific preference for the cost of errors made on different fractions of the data. We propose a simple visualization tool indicating which models are helpful depending on the user's expectations. The usefulness of such an analysis is demonstrated using a selected benchmark analysis as an example.

\section{Motivations}

This section focuses on selected properties of the metrics while realizing that this section aims not to provide an exhaustive overview of the metrics but to indicate the motivation for developing a method that allows fair comparison of imbalanced data classifiers.

\begin{table}[!t]   
\caption{Confusion matrix for a two-class problem.}\vspace{.5em}
\label{table1}
\centering
\begin{tabular}{|cc|c|c|}
  \cline{3-4}
 \multicolumn{2}{c|}{\textsc{ }} & \multicolumn{2}{c|}{\textsc{predicted class}} \\
 \multicolumn{2}{c|}{\textsc{ }} & \textsc{positive} & \textsc{negative}\\
  \hline
    \multirow{6}{*}{\rotatebox[origin=c]{90}{\textsc{true class}}}  &&&\\
    & \textsc{positive} & \emph{True Positive ($TP$)} & \emph{False Negative ($FN$)}\\
   
    &&&\\
    \cline{2-4}
     &&&\\
   & \textsc{negative} & \emph{False Positive ($FP$)} & \emph{True Negative ($TN$)}\\
   &&&\\
    
  \hline
\end{tabular}
\end{table}

Many metrics for classifier evaluation have been proposed  \cite{Japkowicz:2011}. Usually, they are calculated on the basis of \emph{confusion matrix}, 
which summarizes the number of instances from each class classified correctly or incorrectly as the remaining classes \cite{Devijver:1982}. For a two-class classification task, let us consider $2\times2$ confusion matrix (see Tab.~\ref{table1}).

The most popular metric is the \emph{Accuracy} ($Acc$):

\begin{align}
Acc = \frac{TP+TN}{TP+FN+TN+FP}
\end{align}

However, it is easy to show that $Acc$ is heavily biased towards the majority class and could lead to misleading conclusions. It is a good metric only for tasks that use the so-called $0-1$ loss function, i.e., when the cost of errors committed on each class is the same. This observation has led the imbalanced data science community to look for other metrics.  

Often we are interested in classifier evaluation on only a part of the data, i.e, positive or negative data. \emph{True Positive Rate} (\emph{TPR}) also known as \emph{Recall} or \emph{Sensitivity}), 
\emph{True Negative Rate} (\emph{TNR}) known as \emph{Specificity}, and \emph{Positive Predictive Value} (\emph{PPV}) also called \emph{Precision}: 

\begin{align}
&TPR = \frac{TP}{TP+FN}, \\
&TNR = \frac{TN}{TN+FP}, \\
&PPV = \frac{TP}{TP+FP}
\end{align}

It is well known that people do not like to make comparisons based on multiple criteria (metrics). Moreover, solving such a problem leads to multi-criteria optimization, which may return not one but several solutions, so-called non-dominated solutions, i.e., solutions for which none of the metrics can be improved without degrading some of the other criteria. Usually, to select a solution that suits the user's preferences, a multiple criteria decision analysis (MCDA) is employed \cite{Cinelli:2020}. However, that is a difficult task that has been researched for years. MCDA solutions focus on designing a decision-making process to assist the user in deciding. Such a decision-making process primarily helps identify the user's preferences, which could be used in a decision model. An example of such a process is PROMETHEE, which relies on pairwise comparisons to rank alternatives evaluated on multiple criteria \cite{Mareschal:2005}.

An alternative approach is to reduce all criteria to a single one, a function of all used criteria, and select a solution according to its value. This approach has been widely accepted in the scientific community and focuses on imbalanced data classification \cite{Luque:2019}.

A number of metrics have been proposed, generally aggregating \emph{TPR} and \emph{TNR}, or \emph{TPR} and \emph{PPV}. Examples of such metrics are the arithmetic, geometric or harmonic means between the two components: \emph{recall} and \emph{specificity}. There are also other proposals trying to enhance one of the two components of the mean, for example \emph{Index of Balanced Accuracy} \cite{bib07} or $F_{\beta}score$ \cite{bib14}. This work focuses on using $F_{\beta}$, thus let us present its definition

\begin{align}
F_{\beta} = \frac{(\beta^2+1) \times {PPV} \times {TPR} } {\beta^2 \times {PPV} + {TPR} }
\end{align}

The $\beta$ parameter expresses the trade-off between selected simple metrics. It will express how much more critical $TPR$ is to the user than $PPV$. Improper selection of the value of the mentioned parameter can lead to the choice of an inappropriate classifier, e.g., favor the majority class for imbalanced data classification task \cite{Brzezinski:2018}.

Interestingly, many practical recommendations for quality metrics note the importance of the $\beta$ parameter and suggest using indicators for several values (usually 0.5, 1, and 2). Unfortunately, papers comparing the quality of classifiers of imbalanced data typically provide only the $F_1$ metric.

\section{$F_\beta$-plot analysis}

The idea of $F_\beta$-plot is to visualize the $F_{\beta}$ values for each of the analyzed classifiers depending on the $\beta$ value then determine the $\beta$ ranges to indicate which classifier takes the best values.
 
The ranking of the evaluated methods will vary with the value of the $\beta$ parameter. Classifiers with a preference for the majority class (and, therefore, achieving a higher precision value) will be superior as the value of the $\beta$ parameter decreases. While the $\beta$ is converging to $0$, the $F_{\beta}$ value will approach the $precision$ score of the classifier. Similarly, the value of the $F_{\beta}$ will approximate the $TPR$ component, with a higher value of the $\beta$ parameter increasing to infinity. The point of balance between the components is 1 – the value for which the $F_{\beta}$ function is equivalent to the harmonic mean.

\begin{figure}
    \centering
    \includegraphics[width=\textwidth, trim={0em 0 0em 0}, clip]{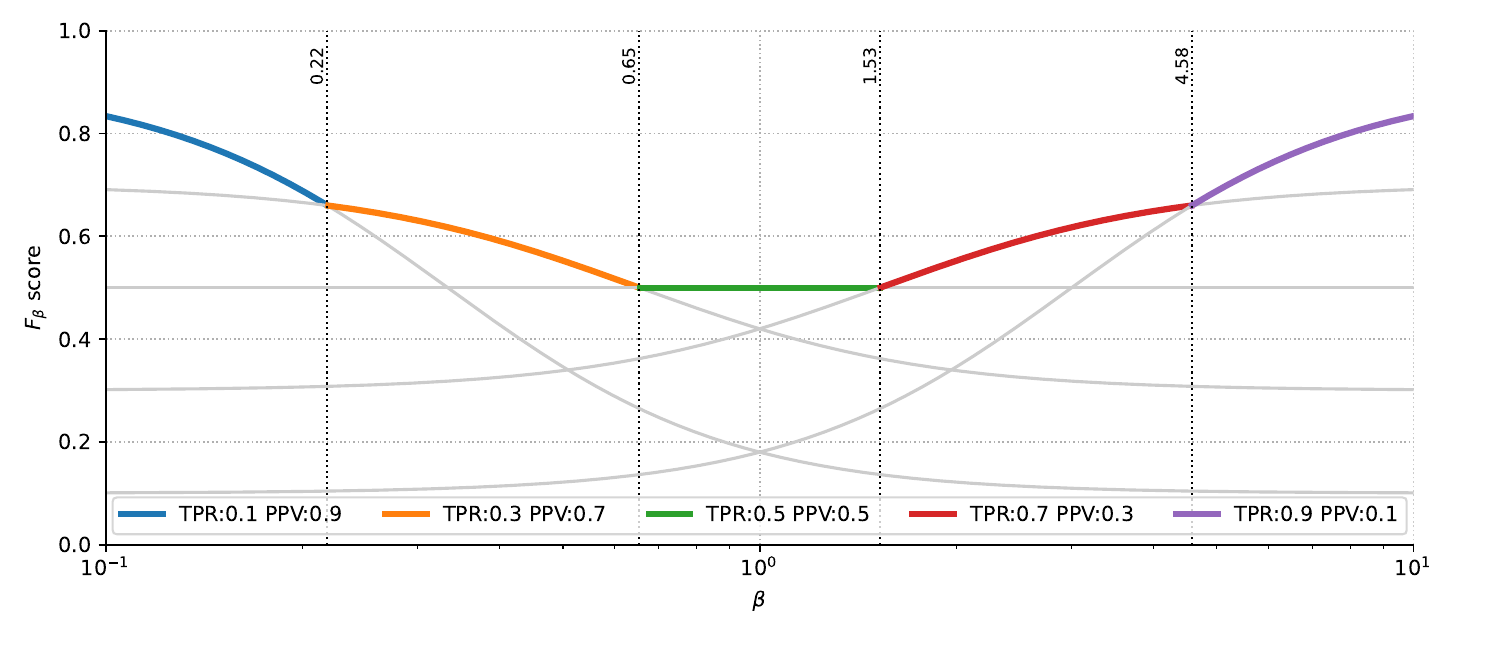}
    \caption{Example of relations between $F_\beta$ and $\beta$ }
    \label{fig:f-relation}
\end{figure}

$F_\beta$-plot is a tool for observing changes in ranking depending on the $\beta$ parameter. A simulation of such a relationship is shown in Figure~\ref{fig:f-relation}. The $\beta$ values are presented on a logarithmic scale to make the results more readable.

Five different scenarios considering different configurations of $TPR$ and $PPV$ were simulated.
As one can observe, it is affirmed that for values of $\beta=1$, a perfectly balanced classifier is preferred. However, even with slight differences in the $\beta$ parameter's value, the curves' ranking changes significantly, preferring results more biased towards one of the parameters. Biased solutions are preferred as well for strongly deviating $\beta$ values.

Noteworthy is the fact that $\beta$ values where curves are intersecting are easily determined

\begin{align}
    \frac{(1 + \beta^2) \cdot \mathrm{PPV}_A \cdot \mathrm{TPR}_A}{(\beta^2 \cdot \mathrm{PPV}_A) + \mathrm{TPR}_A} = \frac{(1 + \beta^2) \cdot \mathrm{PPV}_B \cdot \mathrm{TPR}_B}{(\beta^2 \cdot \mathrm{PPV}_B) + \mathrm{TPR}_B}
\end{align}

\begin{align}
    \beta = \sqrt{\frac{(\mathrm{TPR}_A \cdot \mathrm{TPR}_B \cdot (\mathrm{PPV}_B - \mathrm{PPV}_A)}{(\mathrm{PPV}_A \cdot \mathrm{PPV}_B \cdot (\mathrm{TPR}_A - \mathrm{TPR}_B)}}
\end{align}

\noindent where indexes A and B refer to exemplary classifiers.

Nevertheless, the representation of the achieved scores in their entire range, along with an indication of the intersections (thus, ranking changes), allows a more thorough evaluation of the quality of tested models. At the same time, the $F_{\beta}$-plot provides guidelines to the system designer -- indicating the method that achieves the best performance according to the preference of the system's end user, which should determine the proportion between the cost of errors.

To give a practical example of $F_{\beta}$-plot application, we conducted a simple experiment on a well-known imbalanced dataset – \emph{Thyroid Disease} \cite{misc_thyroid_disease_102}. We trained 92 models using $k$-nearest neighbors algorithm ($k$0NN) combined with a various SMOTE-based oversamplers \cite{Kovacs:2019smotev}.

\begin{figure}
    \centering
    \includegraphics[width=\textwidth, trim={0em 0 0em 0}, clip]{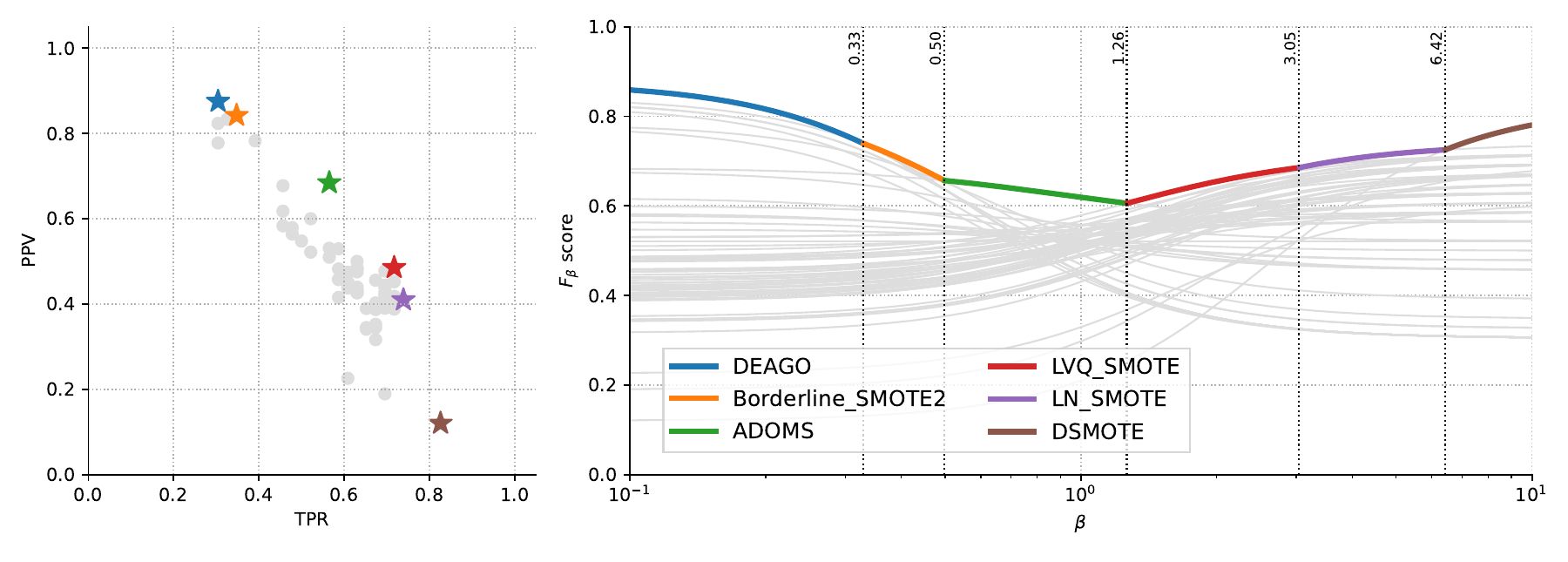}
    \caption{$F_\beta$-plot for \emph{Thyroid Disease} with hold-out evaluation.}
    \label{fig:single-split}
\end{figure}

Firstly, the performance of the models was determined on a single data split with a 20\% hold-out. The $F_\beta$-plot for the obtained results is presented in Figure~\ref{fig:single-split}. A scatter plot was also included, showing the obtained values for $TPR$ -$PPV$ space. The colors mark out the methods that achieved the best result over the analyzed range of $\beta$ values.

We may observe that, depending on the configuration of the $\beta$ parameter, it is possible to distinguish six oversampling methods that result in the best $F_\beta$ values. The balanced method is the \textsc{AHC} algorithm, for which we can also notice it has a slight majority class preference. The other algorithms that prefer the majority class are \textsc{CCR} and \textsc{Gaussian\_SMOTE}, although their curve characteristics are similar and are reflected in the close distance in the scatter plot. From $\beta=1.26$, which we can describe as a slight preference for the minority class, the \textsc{NT\_SMOTE}, \textsc{NRSBoudary\_SMOTE}, and \textsc{ISOMTE}, respectively, start to dominate. We should also point out that the selected models constitute a boundary relative to the point $<1,1>$, which would be considered a \emph{perfect} model.

\begin{figure}
    \centering
    \includegraphics[width=\linewidth, trim={0em 0 0em 0},clip]{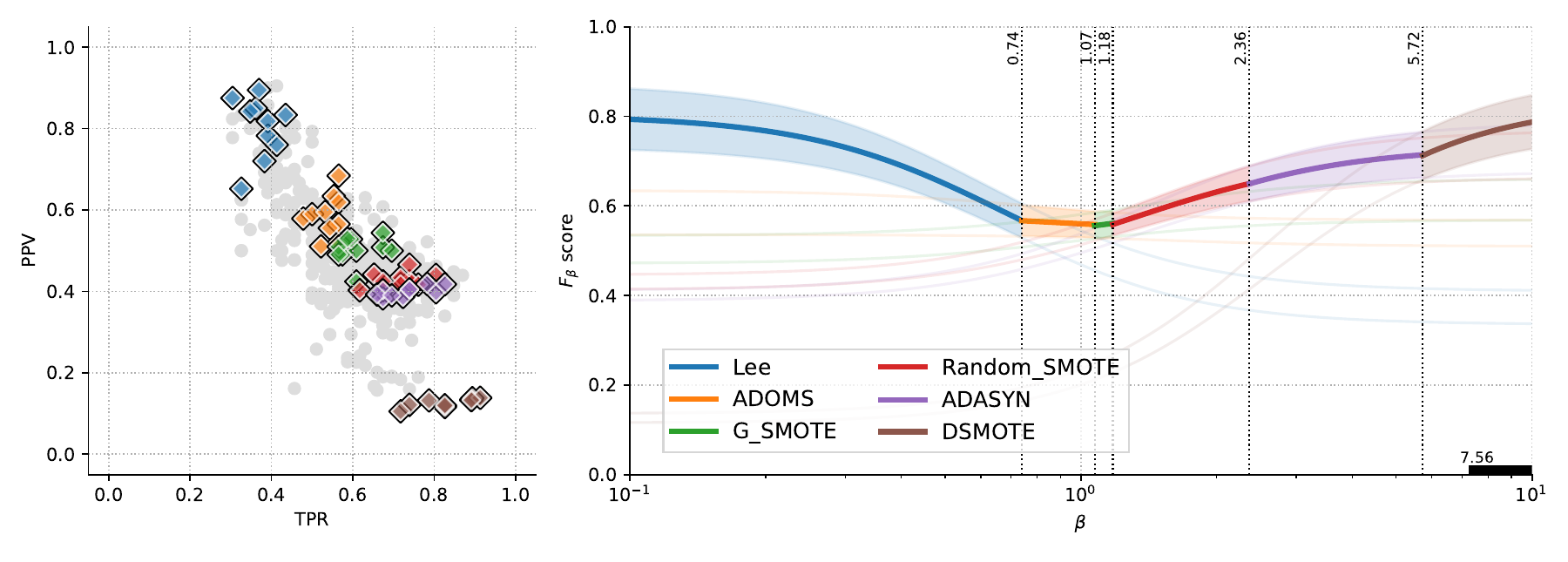}
    \caption{$F_\beta$-plot for \emph{Thyroid Disease} with cross-validation.}
    \label{fig:cv-split}
\end{figure}

Let us notice that many experiments use more rigorous experimental protocols generally based on $k$-fold cross-validation. $F_\beta$-plot could be easily adapted to the new protocol, as shown in Figure~\ref{fig:cv-split}. The plot was expanded to include standard deviations, and the scatter plot marked all pairs of values obtained by methods. The color indicates the methods whose average $F_\beta$ was the highest in the given range.

As expected, the list of selected methods has changed repeated evaluation. The reoccurring algorithms are \textsc{AHC} staying within the range containing $\beta = 1$ and \textsc{ISMOTE} performing better at higher $\beta$ values. The remaining methods have been superseded by others that achieve similar results within the standard deviation range of their $F_\beta$. The observed change reflects using a broad set of resampling algorithms, which tend to generate similar synthetic samples for the learning set. In order to distinguish statistically better algorithms on the sampled interval, a \emph{paired T-test} was performed. The result is presented as a black line near the bottom axis, marking the area where the best algorithm is statistically significantly better than other algorithms. From this, it can be determined that in the case of an extreme preference against recall ($\beta > 7.56$), \emph{ISMOTE} is the only one to achieve the highest score. This information might be practically useful, if the aim of the system designer is to provide a screening test system. In other cases, we have to consider that there is a method similar to the marked one. One should remember that based on the statistical tests performed, it is possible to indicate a group of similar methods, which can also be important information for the system designer.

\section{Experiments}

\begin{figure}
    \centering
    \includegraphics[width=\linewidth, trim={0em 0 0em 0},clip]{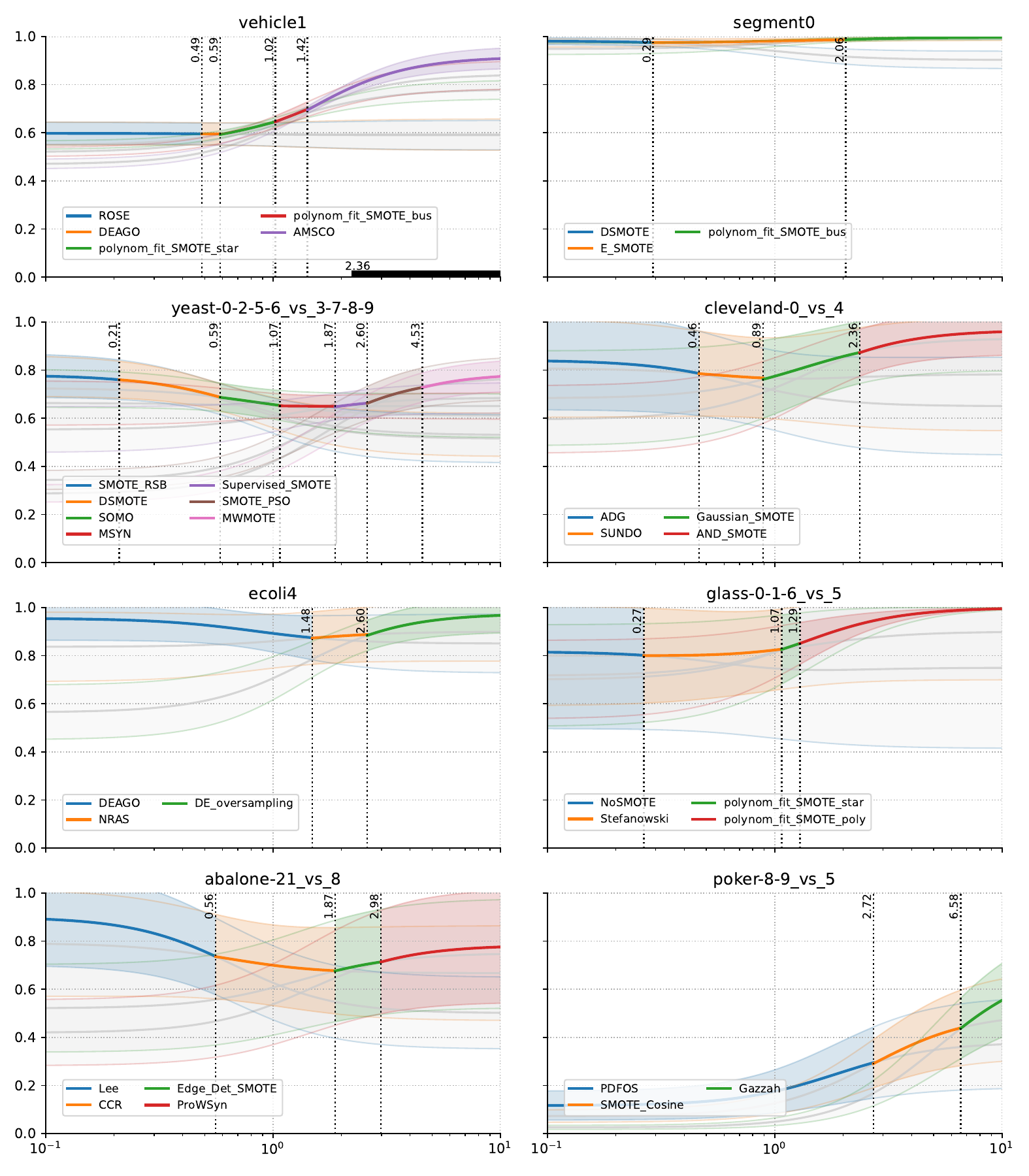}
    \caption{$F_\beta$-plots of selected datasets}
    \label{fig:results}
\end{figure}

The experimental study will present the characteristics of $F_\beta$ plots for selected benchmark imbalanced datasets \cite{derrac2015keel}. A vast pool of oversampling algorithms will be compared \cite{Kovacs:2019comp}, which will be used to preprocess the data, followed by the $k$-NN classifier. For the evaluation protocol we choose \emph{2x5-fold~stratified~cross-validation}. Table~\ref{tab:datasets} presents the eight datasets chosen for this experiment. The code for experiments reproduction, as well as $F_\beta$ plot code is publicly available~\footnote{\url{https://github.com/w4k2/fb-plot}}. The experiment aims to investigate observable relationships and discuss their interpretation. The experiment results are presented in Figure~\ref{fig:results}.

\begin{table}[h]
    \centering
    \caption{Main characteristics of the chosen benchmark datasets}
    \renewcommand{\arraystretch}{1.2}%
    \begin{tabularx}{\columnwidth}{lCCC}
        \hline
        Name & Samples & Features & IR\\
        \hline
            vehicle1                   &  846 & 18 &  2.90 \\
            segment0                   & 2308 & 19 &  6.06 \\
            yeast-0-2-5-6\_vs\_3-7-8-9 & 1004 &  8 &  9.14 \\
            cleveland-0\_vs\_4         &  173 & 13 & 12.31 \\
            ecoli4                     &  336 &  7 & 15.80 \\
            glass-0-1-6\_vs\_5         &  184 &  9 & 19.44 \\
            abalone-21\_vs\_8          &  581 &  8 & 40.50 \\
            poker-8-9\_vs\_5           & 2075 & 10 & 82.00 \\
        \hline
    \end{tabularx}
    \label{tab:datasets}
\end{table}

For most datasets, no statistically significant best classifier was observed throughout the analyzed range, and it is only possible to highlight \textsc{AMSCO} on the $\beta > 2.36$ interval for the \emph{vehicle1} dataset. Other methods achieve the best values on the analyzed ranges, but it should be remembered that other oversampling can be identified to obtain similar results.

Additionally, it could be observed that for most problems, the value of the $F_\beta$ remains high, not falling below 0.6. However, the exception is the strongly imbalanced set of \emph{poker-8-9\_vs\_vs\_5}, for which it is difficult to identify a method to achieve a satisfactory result, preferring precision. Such behavior may be related to the pool of preprocessing methods themselves, which - by design - seek to equalize (or outweigh) the model's bias against the minority class. Another factor that may also affect the observed result is the problem's difficulty, which is not always related to the degree of imbalance, but, for example, the number of objects of the borderline class~\cite{Napierala:2012,skryjomski2017influence}. A similar effect is also observed for the sets \emph{vechicle1} and \emph{glass-1-6\_vs\_5}, and the opposite effect is observed for the set \emph{segment0}, where we can almost always indicate the method for which the $F_\beta$ is near 1. As for the other sets, the curve assembled from best algorithms forms a \emph{"V"} shaped curve, where the lowest $F_\beta$ values are achieved in the near surroundings of $\beta=1$ with a slightly higher tendency towards the minority class.

\section{Conclusion}

The issue of selecting suitable metrics for imbalanced data problems remains pertinent. On one hand, it is recommended to avoid aggregated metrics and instead rely on simple metrics for analysis. On the other hand, it should be acknowledged that analyzing multiple criteria simultaneously may be challenging, especially for less experienced users or those who cannot determine the costs of incorrect decisions related to selected data fractions. This paper presents a simple $F_{\beta}-plots$ method to visualize the results of the experiments, allowing simultaneous evaluation of the quality of multiple methods for different values of $\beta$, i.e., the end-user's expectation of the validity of $TPR$ against $PPV$. 
The $F_{\beta}-plots$ method identifies the costs at which a particular method is beneficial.
Additionally, it indicates the tasks for which a given classifier may be suitable, such as values of imbalance ratio that are proportional to costs between simple metrics, as suggested by Brzezinski et al. \cite{Brzezinski:2020}. Such an analysis provides a broader perspective on the quality and scope of the tested classifiers. 

\begin{figure}
    \centering
    \includegraphics[width=0.5\linewidth]{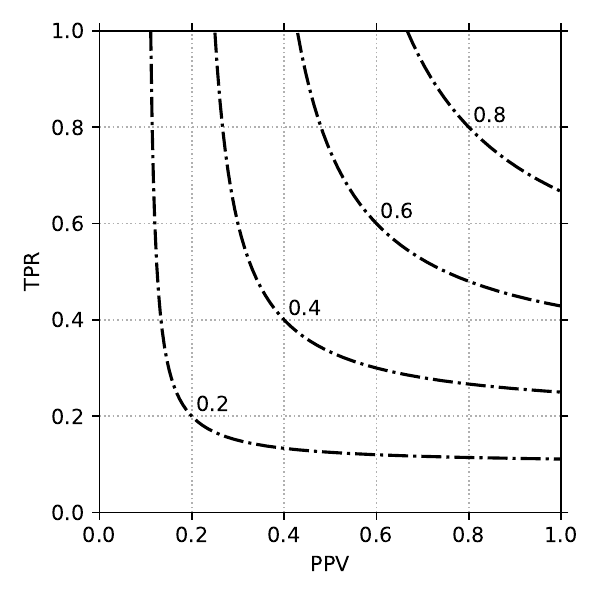}
    \caption{Example of different $F_1$ values in \emph{PPV}-\emph{TPR} space.}
    \label{fig:fbeta}
\end{figure}

However, selecting an aggregate metric can be challenging due to its limited interpretability \cite{Hand:2021} and potential ambiguity, i.e., even if we use the different $\beta$ values, we still face the problem of aggregated metric ambiguousness because the same $F_{\beta}$ value may be taken for different $PPV$ and $TPR$ values (see Figure~\ref{fig:fbeta}). Thus, one might suspect that machine learning methods that use such an aggregated metric as a criterion will be biased toward specific values of simple metrics without providing the information that there are equally good solutions (in terms of a given metric) for other values of \emph{PPV} and \emph{TPR}. Additionally, $F_{\beta}$ ignores the number of true negatives \cite{Christen:2023}. The issues mentioned above were not discussed in this paper and are still waiting to be properly addressed

\subsubsection{Acknowledgements}

This work was supported by the Polish National Science Centre under the grant No. 2019/35/B/ST6/04442.

\bibliographystyle{splncs04}
\bibliography{bibliography}

\end{document}